\documentclass[conference,a4paper]{IEEEtran}
\IEEEoverridecommandlockouts

\usepackage[hidelinks]{hyperref}
\usepackage[cmex10]{amsmath}
\usepackage{amssymb,amsfonts}
\usepackage{dblfloatfix}

\usepackage[ruled,vlined]{algorithm2e}
\usepackage{graphicx}
\graphicspath{{Figures/PDF/}{Figures/PNG/}}

\usepackage{booktabs}
\usepackage{siunitx}
\usepackage[numbers,compress]{natbib}
\usepackage{texnames}
\usepackage{bm,bbm}
\usepackage{orcidlink}
\usepackage{multirow}
\usepackage[table]{xcolor}
\usepackage{makecell}

\begin{document}

\title{\uppercase{DAC-Loop: Distillation-Aware Compression for Remote Sensing CNN}}

\author{
    \IEEEauthorblockN{
        Rico Staedeli \orcidlink{0009-0007-9741-6633}\quad
        Lorena Raichle \orcidlink{0009-0000-4242-2728}\quad
        Konstantinos Tzevelekakis \orcidlink{0009-0000-2502-156X}\quad
        Aron Asefaw \orcidlink{0009-0006-1488-0776}\quad
        Damian Borth \orcidlink{0000-0002-4660-2627}
    }
    \vspace{4mm}
    \IEEEauthorblockA{
        AIML Lab, School of Computer Science, University of St. Gallen
    }
}

\maketitle
\begin{abstract}
While large-scale models achieve strong results in remote sensing scene classification, deploying them on edge devices remains difficult due to limited compute, memory, and energy budgets. Compact CNN backbones are therefore a practical choice in many real-world settings, offering efficient inference with well-optimized implementations on embedded hardware. To push this efficiency further, we investigate how far a knowledge-distilled student model can be compressed while preserving high accuracy. We propose a \underline{D}istillation-\underline{A}ware \underline{C}ompression Loop (DAC-Loop) that integrates
(i) \emph{knowledge distillation} applied both \emph{before} and \emph{after} structured channel pruning, and (ii) an \emph{iterative compression schedule} that alternates structured pruning with unstructured sparsification. Across EuroSAT and NWPU-RESISC45, the DAC-Loop consistently improves the accuracy--compression trade-off relative to standard one-shot pruning and single-stage KD recovery, maintaining strong accuracy even beyond 90\% parameter reduction.
These results demonstrate that combining distillation and pruning in a unified iterative loop offers an effective path toward compact remote sensing models. The Code is available in GitHub. \footnote{\textbf{GitHub Repository:} \href{https://github.com/RicoStaedeli/DAC-Loop}{https://github.com/RicoStaedeli/DAC-Loop}}
\end{abstract}


\section{Introduction}
\label{sec:intro}
\begin{figure}[hbt]
	\centering
	\includegraphics[width=\linewidth]{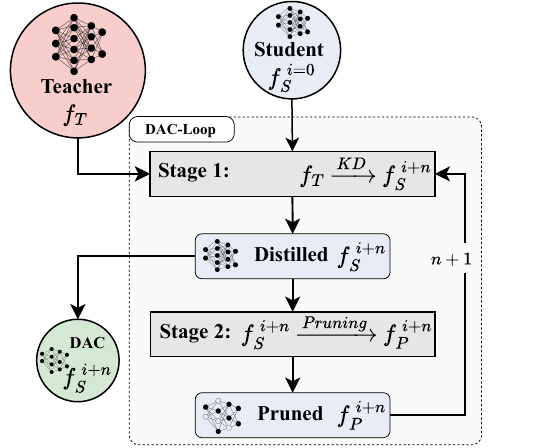}
	\caption{\textbf{Distillation--Aware Compression Loop (DAC-Loop).}
A fixed, pretrained \emph{teacher} network $f_T(\cdot;\theta_T)$ provides supervision for iterative compression of a smaller \emph{student}. Starting from a randomly initialized student $f_S^{\,i=0}$, \textit{Stage 1} performs KD from $f_T$ to obtain a distilled student $f_S^{\,i+n}$. The resulting model is then compressed in \textit{Stage 2} via structured or unstructured pruning, producing a pruned model $f_P^{\,i+1}$. Since pruning typically degrades accuracy, the pruned network subsequently restores performance using the same KD procedure employed in Stage 1. The distilled-and-recovered student can either be used as the final DAC model or re-enter the loop for additional pruning cycles, enabling up to $N$ iterations. Details in Chapter~\ref{dac-detail}.}
    \label{fig:dac_loop}
\end{figure}

Remote Sensing (RS) has benefited greatly from advances in deep neural networks, enabling highly accurate land-use and land-cover (LULC) classification on large-scale datasets such as EuroSAT and NWPU-RESISC45 \cite{xiong_earthnets_2024}. LULC-classification became a central task for monitoring, environmental analysis, and constrained devices decision-making. However, deploying such models on resource-constrained devices remains challenging as state-of-the-art CNN-backbones require substantial memory, compute, and power \cite{zhao_lightweight_2025}. This challenge motivates the development of compact yet accurate RS classifiers. While lightweight architectures exist, they rely on custom, task-specific designs and extensive architecture engineering \cite{liu_efficient_2025}. A promising direction to mitigate this gap lies in teacher-student knowledge distillation (KD), where a high-capacity teacher model guides a smaller student model in training to achieve comparable accuracy \cite{song_symmetrical_2025} \cite{he_remote_2025} \cite{himeur_applications_2024}. Yet, even smaller models such as distilled students can exceed the tight memory and compute limits of resource-constrained platforms \cite{shuvo_efficient_2023}.
The goal of this paper is to explore how much a knowledge-distilled student model can be further compressed while preserving high level of accuracy on RS scene classification. We therefore consider compression as an iterative process in combination with structured and unstructured pruning. In particular, our main contributions are:
\begin{itemize}
    \item We introduce the DAC-Loop (Fig.~\ref{fig:dac_loop}), an iterative distillation-aware compression protocol for CNN backbones that alternates pruning and KD-based recovery under a fixed teacher.
    \item We analyze distillation around structured pruning (KD-1/KD-2) and a pruning schedule that alternates structured and unstructured pruning across iterations.
    \item On EuroSAT and NWPU-RESISC45, ablations and accuracy--compression curves show that the DAC-Loop improves the accuracy--efficiency trade-off over standard one-shot pruning (Fig.~\ref{fig:DAC-comparison}).
\end{itemize}

\noindent The DAC-Loop contributes a \emph{compression protocol} which specifies where distillation is applied, how compression is staged across multiple cycles, and how structured and unstructured pruning are combined.

\section{Related Work}
\label{sec:related}
To enable deployment under tight compute and memory budgets, prior work in RS has explored lightweight architectures and compression of standard CNN backbones \cite{liu_efficient_2025, zhao_lightweight_2025}.
Model compression via pruning and quantization typically exposes an accuracy--efficiency trade-off. Stronger reductions in parameters often lead to measurable performance drops \cite{shinde_model_2025, zou_remotetrimmer_2024}.
For instance, Shinde et al.\ \cite{shinde_model_2025} prune and quantize CNNs on EuroSAT and report substantial efficiency gains with limited accuracy degradation.

Teacher--student KD has been widely adopted to improve compact RS classifiers, frequently enabling small students to approach the accuracy of larger teachers \cite{song_symmetrical_2025, he_remote_2025, himeur_applications_2024}.
However, KD is commonly used as a training technique for fixed student architectures, rather than as a mechanism to support \emph{progressive} model compression.
Outside RS, several prior studies use KD to recover accuracy after pruning. Aghli et al.\ \cite{aghliCombiningWeightPruning2021} combine pruning and KD by using a pruned model in the teacher--student setup, and Muralidharan et al.\ \cite{muralidharan_compact_2024} show that iterative structured pruning with KD-based retraining can yield compact yet accurate models. In RS scene classification, KD and pruning are still most often evaluated separately or in one-shot combinations \cite{song_symmetrical_2025, shinde_model_2025, zou_remotetrimmer_2024}. In contrast, DAC-Loop studies \emph{iterative} RS compression with (i) \emph{KD} applied before and after structured channel pruning and (ii) a scheduled alternation of structured and unstructured pruning within a repeatable protocol.

\section{Methodology}
\subsection{Knowledge Distillation}
\label{sec:KD}

In the first KD stage the predictive knowledge of a teacher model is transferred to a smaller student. Following Hinton et al. \cite{hinton_distilling_2015}, the student is trained using a hybrid distillation objective. Let $T_{\phi}$ and $S_{\theta}$ denote the teacher and student networks, respectively. The KD objective is defined as
\begin{equation}
    L = (1 - \alpha) \, L_{\mathrm{CE}}(y, S_{\theta}(x)) + \alpha \, L_{\mathrm{KD}}(T_{\phi}(x), S_{\theta}(x)),
\end{equation}
where $\alpha$ controls the trade-off between supervised learning and teacher guidance. To select the model individual best suiting KD method, four different KD-strategies were implemented. These methods differ in the definition of $L_{\mathrm{KD}}$. 
A special case is the \textit{KD via Cosine Embedding Loss} variant. For this method the \emph{teacher} and \emph{student} feature representations must have matching dimensionality, since the cosine loss is defined on pairs of vectors in the same space. 
Let $\mathbf{f}^{(t)} \in \mathbb{R}^{d_t}$ and $\mathbf{f}^{(s)} \in \mathbb{R}^{d_s}$ denote the intermediate feature vectors of the teacher and student, respectively. To enable a well-defined similarity computation when $d_t \neq d_s$, a learnable projection layer was introduced which maps the student features into the teacher's feature space. 
The resulting \textit{cosine-embedding KD loss} is then computed between $\mathbf{f}^{(t)}$ and $\mathbf{f}^{(s)}_{\mathrm{proj}}$, ensuring shape compatibility while still allowing architectural flexibility between teacher and student models.
A hyperparameter search is conducted to identify the best-performing individual strategies for each model, which are then used in subsequent stages of the DAC loop.

\subsection{Pruning}
\label{sec:pruning}

To reduce the parameter count, we apply two complementary strategies: (1) global unstructured (weight-level) pruning and (2) structured channel pruning for convolutional layers.

\subsubsection{Global Unstructured Pruning}
\label{unstructured-pruning}
Let $\{\mathbf{W}_\ell\}_{\ell=1}^{L}$ be the weight tensors of all prunable layers (convolutional and linear). We associate binary masks $\{\mathbf{M}_\ell\}_{\ell=1}^{L}$ of identical shape and define masked weights
\begin{equation}
    \widetilde{\mathbf{W}}_\ell = \mathbf{M}_\ell \odot \mathbf{W}_\ell,
\end{equation}
where $\odot$ denotes elementwise multiplication and $(\mathbf{M}_\ell)_i \in \{0,1\}$ indicates whether parameter $i$ is kept ($1$) or pruned ($0$). For a target global sparsity $\alpha \in [0,1)$ (fraction of pruned weights across all prunable layers), we form the multiset $\mathcal{W}$ of all scalar weights. In magnitude pruning, we choose a global threshold $\tau$ such that a fraction $\alpha$ of $w\in\mathcal{W}$ satisfies $|w|\le \tau$, and set
\begin{equation}
    (\mathbf{M}_\ell)_i =
    \begin{cases}
        0, & \text{if } \bigl|\bigl(\mathbf{W}_\ell\bigr)_i\bigr| \leq \tau, \\[2pt]
        1, & \text{otherwise.}
    \end{cases}
\end{equation}
Masks are applied via re-parameterization during training so pruned weights do not contribute to the forward pass.

\subsubsection{Structured Channel Pruning}
\label{structured-pruning}
While global unstructured pruning increases weight-level sparsity, it does not change the layer shapes and thus does not directly reduce the number of channels. To obtain a structurally smaller network, we additionally perform structured pruning of convolutional channels following the channel-pruning formulations introduced by Li et al. \cite{li_pruning_2017} and He et al. \cite{he_channel_2017}.
Consider a convolutional layer $\ell$ with weight tensor
\begin{equation}
    \mathbf{W}_\ell \in \mathbb{R}^{C_{\text{out}} \times C_{\text{in}} \times k_h \times k_w},
\end{equation}
where each slice $\mathbf{W}_{\ell,i,:,:,:}$ corresponds to the kernel associated with output channel $i \in \{1, \dots, C_{\text{out}}\}$. We define the importance score of channel $i$ as the $\ell_2$ norm of its kernel:
\begin{equation}
    s_{\ell,i} = \left\| \mathbf{W}_{\ell,i,:,:,:} \right\|_2^2
    = \sum_{c=1}^{C_{\text{in}}} \sum_{u=1}^{k_h} \sum_{v=1}^{k_w}
      \bigl( \mathbf{W}_{\ell,i,c,u,v} \bigr)^2.
\end{equation}
For a defined pruning ratio $r_\ell \in [0,1)$ for layer $\ell$, we compute $n_\ell = \lfloor r_\ell C_{\text{out}} \rfloor$ and prune the $n_\ell$ output channels with the smallest scores $s_{\ell,i}$.
In practice, removing channels from one layer affects the input dimensionality of subsequent layers. To ensure architectural consistency, a dependency graph of the network is constructed. Pruning is then applied to the entire group atomically, guaranteeing that all tensor shapes remain compatible and the resulting network is a valid, smaller architecture. 

\begin{table*}[hbt]
    \centering
    \caption{\textbf{KD-1 \& KD-2} both represent \textbf{Stage 1 Training with KD} in our DAC-Loop (Figure \ref{fig:dac_loop}) and indicate whether KD is applied pre (KD-1) or post (KD-2) pruning. \textbf{N} denotes the number of DAC-Loop iterations.}
    \setlength{\tabcolsep}{5pt}
    \begin{tabular}{c l c c c r r c r r}
        \hline
        \multicolumn{10}{c}{\textbf{Ablation Study Results for DAC-Loop with a ResNet18}} \\[2pt]
        \hline
         & \textbf{Model} & \textbf{KD-1 } & \textbf{KD-2} & \textbf{N}
         & \textbf{Compression} & \textbf{Parameters} & \textbf{Prun. Type}
         & \textbf{Val. Acc. (EuroSAT)} & \textbf{Val. Acc. (RESISC45)} \\
        \hline

        \multirow{3}{*}{\rotatebox{90}{KD}}%
         & ResNetTiny   & \checkmark & --           & 1 & 0\% & 1.0 M  & -- & 96.49{\scriptsize $\pm$ 0.17} & 82.63{\scriptsize $\pm$ 0.32}\\
         & ResNet10mini & \checkmark & --           & 1 & 0\% & 2.0 M  & -- & 97.11{\scriptsize $\pm$ 0.15} & 86.36{\scriptsize $\pm$ 0.37}\\
         & ResNet18     & \checkmark & --           & 1 & 0\% & 11.2 M & -- & 98.61{\scriptsize $\pm$ 0.08} & 91.95{\scriptsize $\pm$ 0.17}\\
        \hline
        
        \multirow{3}{*}{\rotatebox{90}{Unstr.}}%
         & ResNet18 & \checkmark & --           & 1 & 0\% & 11.2 M  & Unstructured & 98.58{\scriptsize $\pm$ 0.06} & 92.59{\scriptsize $\pm$ 0.02}\\
         & ResNet18 & --         & \checkmark   & 1 & 0\% & 11.2 M  & Unstructured & 98.41{\scriptsize $\pm$ 0.02} & 92.27{\scriptsize $\pm$ 0.20}\\
         & ResNet18 & \checkmark &\checkmark    & 1 & 0\% & 11.2 M  & Unstructured & 98.67{\scriptsize $\pm$ 0.05} & 92.22{\scriptsize $\pm$ 0.03}\\
        \hline

        \multirow{4}{*}{\rotatebox{90}{Comp.\ +}}%
         & ResNet18 & --         & --         & 1 & 46.0\% & 6.0 M & Structured & 98.24{\scriptsize $\pm$ 0.02} & 90.78{\scriptsize $\pm$ 0.11}\\
         & ResNet18 & \checkmark & --         & 1 & 46.0\% & 6.0 M & Structured & 98.37{\scriptsize $\pm$ 0.06} & 82.02{\scriptsize $\pm$ 0.15}\\
         & ResNet18 & --         & \checkmark & 1 & 46.0\% & 6.0 M & Structured & 98.46{\scriptsize $\pm$ 0.04} & 92.10{\scriptsize $\pm$ 0.04}\\
         & \cellcolor{gray!10}ResNet18 & \cellcolor{gray!10}\checkmark & \cellcolor{gray!10}\checkmark & \cellcolor{gray!10}1
           & \cellcolor{gray!10}46.0\% & \cellcolor{gray!10}6.0 M &  \cellcolor{gray!10}Structured
           & \cellcolor{gray!10}98.48{\scriptsize $\pm$ 0.06} & \cellcolor{gray!10}92.22{\scriptsize $\pm$ 0.03}\\
        \hline

        \multirow{4}{*}{\rotatebox{90}{Comp.\ ++}}%
         & ResNet18& --         & --         & 1 & 75.0\% & 2.8 M & Structured & 97.82{\scriptsize $\pm$ 0.10} & 87.86{\scriptsize $\pm$ 0.17}\\
         & ResNet18& \checkmark & --         & 1 & 75.0\% & 2.8 M & Structured & 98.20{\scriptsize $\pm$ 0.04} & 89.98{\scriptsize $\pm$ 0.29}\\
         & ResNet18& --         & \checkmark & 1 & 75.0\% & 2.8 M & Structured & 98.09{\scriptsize $\pm$ 0.06} & 90.13{\scriptsize $\pm$ 0.20}\\
         & \cellcolor{gray!10}ResNet18& \cellcolor{gray!10}\checkmark & \cellcolor{gray!10}\checkmark & \cellcolor{gray!10}1
           & \cellcolor{gray!10}75.0\% & \cellcolor{gray!10}2.8 M & \cellcolor{gray!10}Structured
           & \cellcolor{gray!10}98.41{\scriptsize $\pm$ 0.16} & \cellcolor{gray!10}91.11{\scriptsize $\pm$ 0.08}\\
        \hline

        \multirow{2}{*}{\rotatebox{90}{Loop}}%
         & \cellcolor{gray!10}ResNet18 & \cellcolor{gray!10}\checkmark & \cellcolor{gray!10}\checkmark & \cellcolor{gray!10}7
           & \cellcolor{gray!10}84.7\% & \cellcolor{gray!10}1.7 M & \cellcolor{gray!10}both
           & \cellcolor{gray!10}\textbf{97.63}{\scriptsize $\pm$ 0.11} & \cellcolor{gray!10}\textbf{91.61}{\scriptsize $\pm$ 0.24}\\
         & ResNet18 & \checkmark & \checkmark & 9 & 91.7\% & 0.9 M & both
           & 96.97{\scriptsize $\pm$ 0.46} & 90.64{\scriptsize $\pm$ 0.15}\\
        \hline
    \end{tabular}
    \label{tab:r18-ablation-combined}
\end{table*}

\subsection{Distillation–Aware Compression Loop (DAC-Loop)}
\label{dac-detail}
As illustrated in Fig.~\ref{fig:dac_loop}, a \emph{teacher} model \(f_T(\cdot;\theta_T)\) is employed, instantiated as a ResNet50. The model $f_T$ is pretrained once and kept frozen throughout the entire DAC-Loop, therefore the same teacher is reused at every iteration. A smaller \emph{student} network \(f_S^{\,i+n}(\cdot;\theta_S)\), in this work a ResNet18, is randomly initialized and trained with the defined KD-method and the pretrained \emph{teacher} model $f_T$ in \textit{Stage 1}. The distilled student model $ f_S^{\,i+n}$ obtained from Stage 1 is then passed to \textit{Stage 2}, where it is compressed using structured or unstructured pruning yielding the model $ f_P^{\,i+1}$. This requires a subsequent fine-tuning step to recover performance, which is again performed using the KD-based training procedure from Stage 1, where the compressed model continues to benefit from the same teacher’s guidance. After pruning and KD recovery, the student $f_S^{\,i+n}$ is either used as output DAC-Model or enters again Stage 2, where it is pruned again. This loop allows the process to run for \textit{N} compression cycles. To support multiple DAC iterations, we employ a compression scheduler that assigns specific pruning strategies, hyperparameter to each loop cycle. This enables different compression across iterations, e.g. unstructured pruning in the first cycle and structured pruning in the next.

\section{Experiments \& Results}
\label{sec:experiments}

\subsection{Data \& Setup}
To evaluate the proposed DAC-Loop the RS datasets \textit{EuroSAT} \cite{helber_eurosat_2019} and \textit{NWPU-RESISC45} \cite{cheng_remote_2017} were used. To train the models, we adopt a 60-20-20 split. Normalization is applied across all experiments, while data augmentation is employed only for training the ResNet50 teacher model. All experiments were conducted on a Tesla V100-SXM3-32GB. We trained all models using the AdamW optimizer with an initial learning rate of $3 \times 10^{-4}$ and a batch size of 256. Unless stated otherwise, the learning rate was scheduled using a cosine annealing schedule over the full 50 training epochs.

\subsection{Baselines}
\label{sec:Baselines}
Table \ref{tab:baselines-eurosat-resisc} reports the model baselines we trained from scratch and compares them to published baselines in prior work \cite{sharma_improving_2025, neumann_-domain_2019, helber_eurosat_2019}. The results show that our implementations reach comparable accuracy on both EuroSAT and NWPU-RESISC45, providing a reliable reference point for the compression experiments that follow. All our experiments are conducted with four different seeds to assess the robustness of the results. Except for the ResNet50 teacher model, all baselines were trained without data augmentation, while the teacher was trained with data augmentation for 100 epochs.

\begin{table}[hbt]
    \centering
    \caption{Baseline validation accuracy}
    \begin{tabular}{lrll}
        \toprule
        \textbf{Model} & \textbf{Param} & \textbf{EuroSAT} & \textbf{RESISC45} \\
        \midrule
        
        \multicolumn{3}{l}{\textit{Baselines - others}} \\
        ResNet18 \cite{sharma_improving_2025} & 11.7 M & 98.06 & 92.10 \\
        ResNet50 \cite{neumann_-domain_2019}  & 24.0 M & 99.20 & 96.80 \\
        ResNet50 \cite{helber_eurosat_2019}   & 24.0 M & 98.57 & -- \\
        
        \midrule
        
        \multicolumn{3}{l}{\textit{Baselines - ours}} \\
        ResNetTiny     & 0.9 M  & 95.77{\scriptsize $\pm$ 0.06} & 80.15{\scriptsize $\pm$ 0.54}\\
        ResNet10mini   & 2.0 M  & 96.41{\scriptsize $\pm$ 0.09} & 83.43{\scriptsize $\pm$ 0.44}\\
        ResNet10       & 4.9 M  & 98.37{\scriptsize $\pm$ 0.05} & 88.33{\scriptsize $\pm$ 0.19}\\
        ResNet18       & 11.2 M & 98.34{\scriptsize $\pm$ 0.05} & 89.76{\scriptsize $\pm$ 0.14}\\
        ResNet50 - Teacher       & 24.6 M & 96.79{\scriptsize $\pm$ 0.18} & 92.44{\scriptsize $\pm$ 0.15}\\
        \bottomrule
    \end{tabular}
    \label{tab:baselines-eurosat-resisc}
\end{table}

\subsection{Ablation Study}
\label{sec:ablationstudy}
In Table~\ref{tab:r18-ablation-combined}, the results of our DAC-Loop ablation study are grouped into five blocks: \emph{KD} (no compression, only KD), \emph{Comp.+} (moderate structured compression strength at 46.0\% parameter reduction), \emph{Unstr.} (unstructured compression), \emph{Comp.++} (stronger structured compression strength at 75.0\%), and \emph{Loop} (multiple DAC-Loop iterations). Within each block, we systematically test the KD components or vary the number of iterations $N$. Across both one-shot structured compression degrees (\emph{Comp.+} and \emph{Comp.++}), the grey-highlighted rows show that using KD in both stages of our DAC-Loop yields the best performance. This confirms that integrating pre-pruning KD with post-pruning KD in structured channel pruning is more effective than either approach alone. The \emph{Loop} experiments further illustrate the benefit of iterating the DAC-Loop: with $N=7$ iterations we obtain a model compressed to 1.7M parameters that still achieves an accuracy of 91.61\% on the NWPU-RESISC45 Dataset (bold in Table~\ref{tab:r18-ablation-combined}). This ablation study demonstrates that repeated DAC-Loop cycles enable progressively higher compression while preserving, and even enhancing, model accuracy. It can also be observed that the results on both the NWPU-RESISC45 and EuroSAT datasets consistently confirm this finding. However, the improvement obtained with DAC-Loop is substantially more pronounced on NWPU-RESISC45. \\

Table~\ref{tab:r18-ablation-combined} shows that applying global unstructured magnitude pruning boosts accuracy to 92.22\%, outperforming only using KD. Because unstructured pruning removes individual weights rather than whole channels, it leaves the architecture and parameter count unchanged. While this sparsity doesn’t compress the model, it helped boost accuracy beyond what structured pruning alone achieves. Our empirical results show that the DAC-Loop achieves the best performance when both complementary pruning methods are integrated into a unified framework. In the ablation study, the compression results obtained with DAC-Loop (last two rows of Table~\ref{tab:r18-ablation-combined}) are produced using the pruning schedule summarized in Table~\ref{tab:pruning_schedule}.

\begin{table}[hbt]
    \centering
    \caption{Pruning schedule parameters and validation accuracy for ResNet18 on NWPU-RESISC45}
    \label{tab:pruning_schedule}
    \begin{tabular}{clrl}
    \toprule
    \textbf{Iteration (N)} & \textbf{Pruning Type} & \textbf{Parameter} & \textbf{Val. Acc.} \\
    \midrule
    1 & Unstructured & 11.2 M & 92.18{\scriptsize $\pm$ 0.06} \\
    2 & Structured   & 6.0 M  & 92.41{\scriptsize $\pm$ 0.11}\\
    3 & Unstructured & 6.0 M  & 92.53{\scriptsize $\pm$ 0.12}\\
    4 & Structured   & 3.2 M  & 92.18{\scriptsize $\pm$ 0.09}\\
    5 & Unstructured & 3.2 M  & 92.21{\scriptsize $\pm$ 0.08}\\
    6 & Structured   & 1.7 M  & 91.54{\scriptsize $\pm$ 0.14}\\
    7 & Unstructured & 1.7 M  & 91.61{\scriptsize $\pm$ 0.24}\\
    8 & Structured   & 0.9 M  & 90.58{\scriptsize $\pm$ 0.11}\\
    9 & Unstructured & 0.9 M  & 90.64{\scriptsize $\pm$ 0.16}\\
    \bottomrule
    \end{tabular}
\end{table}

\subsection{Results and Discussion}
\label{sec:results_discussion}
\begin{figure}[hbt]
  \centering
  \includegraphics[width=\columnwidth]{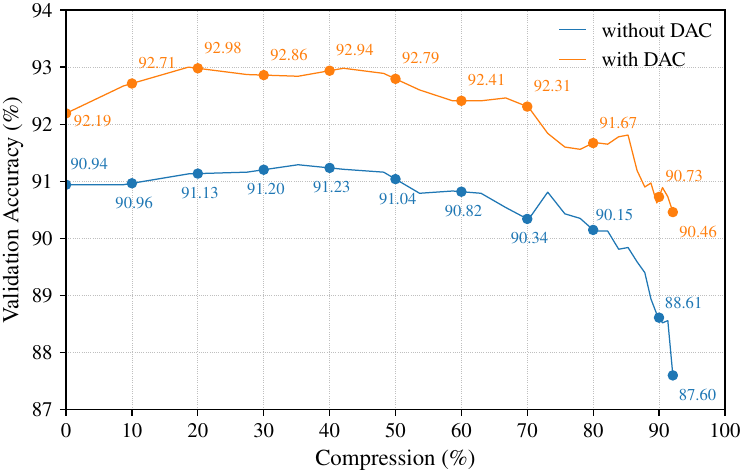}
  \caption{\textbf{Effect of DAC-Loop on the accuracy–compression trade-off}: comparison of validation accuracy on NWPU-RESISC45 under increasing compression for ResNet18 with (orange) and without (blue) the proposed DAC-Loop. Across all compression levels, the DAC-Loop maintains consistently higher accuracy than one-shot pruning.}
  \label{fig:DAC-comparison}
\end{figure}

Figure \ref{fig:DAC-comparison} illustrates the effect of the DAC-Loop on the accuracy–compression trade-off for ResNet18 on NWPU-RESISC45. In this experiment we employ the same ResNet18 with an without the DAC-Loop. Across the entire compression range, the DAC-Loop curve remains consistently above the purely one-shot pruned baseline, demonstrating that distillation-guided recovery stabilizes accuracy. The baseline (blue in Fig. \ref{fig:DAC-comparison}) consist of one-shot compression with varying pruning levels, followed by a subsequent stage of supervised fine-tuning. The difference persists across all compression rates between 10\% and 90\%.  This confirms that the DAC-Loop does not only mitigate pruning loss, it fundamentally shifts the trade-off, enabling significantly smaller models to maintain strong performance. \\

Additionally, Table \ref{tab:inference_comparison} shows that at 256×256 input resolution, the DAC-Loop reduces dense compute from 9.02G to 0.81G MACs (11.16×) and achieves a 4.42× inference speedup on a Tesla V100-SXM3-32GB with a batch size of 256.

\begin{table}[hbt]
\centering
\caption{Efficiency comparison of the full and DAC-Loop–compressed ResNet18 models on the RESISC45 validation set.}
\label{tab:inference_comparison}
\begin{tabular}{lrrl}
\hline
\textbf{Metric} & \textbf{Full} & \textbf{Compressed} & \textbf{Speedup} \\
\hline
Parameters {\small$\downarrow$}              & 11.2 M   & 0.9 M    & $\times 12.44$ \\
FLOPs {\small$\downarrow$}              & 18.05 G   & 1.62 G    & $\times 11.16$ \\
MACs {\small$\downarrow$}               & 9.02 G    & 0.81 G    & $\times 11.16$ \\
Inference (Batch) {\small$\downarrow$}  & 288.25 ms & 65.18 ms  & $\times 4.42$ \\
\hline
\end{tabular}
\end{table}

\section{Conclusion}
\label{sec:conclusion}
We introduced the Distillation-Aware Compression Loop (DAC-Loop), which is a protocol that combines knowledge distillation with structured and unstructured pruning for remote-sensing CNNs. The loop alternates KD-based training and pruning, guided by a compression scheduler that selects pruning type and sparsity per iteration, progressively compressing students while realigning them with the teacher.

On EuroSAT and NWPU-RESISC45, the DAC-Loop yields compact, high-performing ResNet students. For a ResNet18 student on NWPU-RESISC45, it reduces parameters by over 90\% while matching or exceeding the supervised baseline. Ablations show that applying KD before and after structured channel pruning consistently outperforms pure pruning, and that iterative DAC cycles produce highly compressed models that retain strong accuracy. Global unstructured magnitude pruning further improves generalization without altering the architecture, and combining it with structured pruning inside the DAC-Loop improves regularization and true architectural compression. The compression--accuracy curves in Figure~\ref{fig:DAC-comparison} confirm that the DAC-Loop shifts the trade-off toward smaller models, with DAC-compressed networks remaining more accurate than purely pruned counterparts across the full compression range. Overall, our study of iterative pruning in combination with KD shows promising results to generate small yet high performing models for RS applications. 

\newpage
\small
\bibliographystyle{IEEEtranN}
\bibliography{references}

\end{document}